# Planning Courses for Student Success at the American College of Greece

Ioannis T. Christou, Evgenia Vagianou, George Vardoulias

{ichristou, jes, gvardoulias}@acg.edu

Department of Information Technology

Deree College, The American College of Greece

Aghia Paraskevi, Greece

## Abstract

*We model the problem of optimizing the schedule of courses a student at the American College of Greece will need to take to complete their studies. We model all constraints set forth by the institution and the department, so that we guarantee the validity of all produced schedules. We formulate several different objectives to optimize in the resulting schedule, including fastest completion time, course difficulty balance, and so on, with a very important objective our model is capable of capturing being the maximization of the expected student GPA given their performance on passed courses using Machine Learning and Data Mining techniques. All resulting problems are Mixed Integer Linear Programming problems with a number of binary variables that is in the order of the maximum number of terms times the number of courses available for the student to take. The resulting Mathematical Programming problem is always solvable by the GUROBI solver in less than 10 seconds on a modern commercial off-the-self PC, whereas the manual process that was installed before used to take department heads that are designated as student advisors more than one hour of their time for every student and was resulting in sub-optimal schedules as measured by the objectives set forth.*

## 1. Introduction

The American College of Greece (ACG) is the largest American private, non-profit higher-education institute in Europe, originally founded in 1875 in Smyrna, and currently located in Athens, Greece. It consists of three major divisions: (a) the Pierce College for secondary education, (b) the Deree College for undergraduate and graduate education, and (c) the ALBA graduate business school. The Deree College is accredited by the New England Commission of Higher Education (NECHE), and it currently enrolls more than 5,000 undergraduate and graduate students. Deree College has 4 schools (Business & Economics, Liberal Arts & Sciences, Frances Rich School of Fine & Performing Arts, and the Graduate School of Arts & Sciences offering Master's level degrees in applied psychology, communication, leadership, and data science.) In total, the Deree College has 15 departments, offering one or more majors each in their corresponding areas.

Students enrolled at Deree College are encouraged, and do seek advice from their academic advisors in order to plan for their course of action while studying at Deree: during their first year of study, the students usually get advice from the student services office, and after that, they are directed to their major department. From that point on, the department head and/or program coordinator is their main source of advice regarding their planned course schedule for the remainder of their studies. This results in significant pressure to department heads, who, besides their normal teaching, scholarly and administrative duties, are also required to prepare customized course schedules for every student who chooses their department as their major. Besides an initial interview with the student to decide on elective courses that the student would like to take, the department head or program coordinator must come up with a course schedule that obeys a large number of complex constraints (see details in section 2.3) and optimizes a number of objective criteria that are prioritized according to student wishes. Since manual optimization by hand for such a difficult problem is out-of-the-question, the department heads usually resort to predefined templates of feasible course schedules that they then tamper with until they agree with the student's current status (what courses they have passed so far, their GPA etc.) and "look good enough" for the student.

The process described above is a heuristic process that takes up a significant amount of time for the department head or program coordinator; it has been estimated that each student schedule usually takes more than one hour of intense study by the head/coordinator to resolve, which translates to more than 50 hours of work per semester per head/coordinator for what is a pure scheduling optimization problem. Given that there are 15 departments at Deree College, this results in approximately 750 hours of intense human labor per semester, spent by faculty that are by default already high up in the hierarchy.

Given the reasons above, we (all faculty at the Information Technology Department) took up the challenge of automating this process by offering students a program that when fed with current student data and desires, comes up with a feasible schedule obeying all the complex constraints in place for every department of Deree College, and at the same time optimizes a number of objectives according to student desires. The result is the ACG S̲tudent C̲ou̲rse Schedule Optimiz̲er (SCORER), a small (just above 5,000 lines-of-code in Java) desktop program. SCORER accepts as input the following:

- Student data (entered through the program GUI):
    - The courses a student has already passed
    - The desired courses the student wishes to take
    - Student choices regarding summer courses, maximum desired number of courses per term, and maximum desired number of courses to take concurrently when undertaking their thesis (ICT Capstone project)
    - The student's priorities over the objectives implemented in the program
- Program data (read from input configuration files):
    - Course information about each course offered by the college, including course code, title, number of credits, prerequisite and corequisite constraints, terms when the course is offered in the future, course difficulty measured in a scale from 0 to 10 and so on
    - Constraints regarding focus-areas within a major that can be complex subset-selection constraints ("student must take 2 of the following 4, and another 2 of the following 5" and so on)
    - Other constraints information such as maximum number of credits allowed to be registered for in a semester, maximum number of semesters before graduation etc.

SCORER creates a complex Mixed-Integer Linear Program (MILP) (Wolsey & Nemhauser, 1988), (Christou (2011), chapter 1) and calls GUROBI (Gurobi, 2022a), one of the fastest commercial optimization solvers in existence today to solve it. Once GUROBI finds the optimal solution, the problem translates the optimization problem variables back to the problem domain variables, and displays to the student their corresponding schedule of courses in human-readable format. An edit-area allows the user to change the term when they take any given course planned for them, or even to completely reject a proposed course and re-optimize given the new user inputs.

## 1.1. Related Work

There is a large body of work on higher education curricular optimization, see for example (Thompson-Arjona 2019), (Heileman et al, 2019), (Unal & Uysal, 2014) but not on the specific topic of computing individual student course plans. The M.Sc. thesis of Thompson-Arjona in particular, considers some of the constraints put forth by course pre-requisite and co-requisites but only in the context of designing a more balanced curriculum for Electrical Engineering programs that attempts to avoid "toxic course combinations" and bring courses as close as possible to their pre-requisites in terms of when they are offered. The work is not oriented towards optimizing individual student plans. Further, the pre-requisite and co-requisite constraints formulated in the thesis only apply to mandatory courses of the curiculum, not to courses that might be elective courses.

In the commercial world, the Prepler system ([www.prepler.com)](www.prepler.com) does have many of the features that SCORER has and of course has a lot more analytics/dashboard features, but does not explicitly state what the objectives of the produced degree plans are, or even whether it actually optimizes anything behind the scenes.

There is also a rapidly growing body of software for course scheduling that allows students to create their weekly course schedule in the form of calendar applications (see for example Comm and Mathaisel (1988) for an early discussion on class and calendar scheduling software requirements, Miranda (2010) for a more recent study, or the online web-apps at [https://www.kalena.es](https://www.kalena.es), [https://www.roverd.com](https://www.roverd.com) or [https://www.ofcourse.org](https://www.ofcourse.org).) All these applications however are concerned with time-tabling of courses per semester, and not with the long-term planning of courses for an individual student throughout their studies at an institution which is our current focus.

In the rest of the paper, we first present the modeling and formulation of the course scheduling problem at Deree College (section 2), in section 3 we present the details of our program implementation, in section 4 we present some data regarding the efficiencies achieved by the adoption of the system by the "Information Technology" and the "Cyber-security and Networks" programs of the IT Department, and in section 5 we present our conclusions and future directions for our work.

## 2. Problem Formulation

### 2.1 Problem Parameters

Let $\{c_1, \dots c_N\}$ denote the total number of courses available to a student, indexed by the set $C = \{1,2,\dots N\}$. Each course $c_i$ has an associated number of credits $\bar{c}_i$ (often equal to 3, but many courses have different credit value, e.g. courses with labs usually have 4 credits.) We also associate a non-negative difficulty level $\bar{d}_i \geq 0$ with each course $c_i$ that is estimated via the student failure rates in the past 3 years (for courses that have been taught all this time in the past) or via the average letter grade of the student cohorts of the past year (if the course has been taught at least once), or if the course has never been taught before, simply by the instructor's "best guess". The difficulty level is zero for an "average difficulty" course, and can rise to any positive value up to 10, for more difficult courses. The difficulty level of courses plays an important role in the way the optimal schedule is constructed since as we shall see it affects the objective function to optimize under any scenario.

Returning to the set of courses, several important subsets of the set $C$ are as follows:

- $LE \subset C$ is the set of all courses that count as "Liberal Education" courses that must be completed for any degree. A minimum of $L_c$ credits must be taken from liberal education courses for a degree to be granted; this minimum number of credits is currently set to 43.
- $R_j \subset C, j = 1, \ldots k$ are sets of courses, from each of which a student must take at least a certain number of courses $r_j > 0$.
- $L_i \subset C, i = 4,5,6$ are so-called "Level-4", "Level-5", and "Level-6" courses that as we shall see impose certain restrictions on *when* a student can register for them.

There is a minimum number of credits required for graduation, which is denoted by $T_c$ and which is currently set to 121. There is also a maximum allowed number of credits $C_{max}$ a student is allowed to take per semester, which is currently set to 17 cr (which is further increased to 20 cr for honors students.)

Students take courses each year of their studies, during various terms (semesters), which are divided in {Fall, Spring, Summer1, Summer2, SummerTerm} terms. Fall and Spring terms last 13 weeks each, as expected (September to December, and January to April respectively). Summer1 and Summer2 terms are fast-paced intense 4 week long terms that run during June and July respectively, while SummerTerm is a somewhat more relaxed term that lasts 8 weeks starting end of May and running until the last week of July.

The notion of "pre-requisite" and "co-requisite" courses is based on the order by which students progress through their studies, and the knowledge needed by more advanced classes. Unfortunately, the natural order Fall →Spring → Summer1 → Summer2 breaks down when we consider the SummerTerm that runs concurrently with Summer1 and Summer2. To make things clear, if a course X is offered in the Fall semester and another course Y is offered in the Spring semester and lists X as a pre-requisite, then a student who successfully passed X in the Fall, may register for Y in the Spring semester. Similarly, if X is offered in the Spring and the student passes it, then if course Y is offered in Summer1 the student may register for Y in Summer1. However, if X is offered in Summer1 and Y is offered in SummerTerm, then the student cannot register for course X in Summer1 and for course Y in SummerTerm (SummerTerm is not considered "after" Summer1 or Summer2). However, SummerTerm is still considered "after" the Spring semester.

Given the above, we still impose the following order in the academic terms of any particular year, and we shall explain later on how we manage to enforce pre-requisite and co-requisite constraints on all courses scheduled for a student:

Fall=1, Spring=2, Summer1=3, Summer2=4, SummerTerm=5.

We assume there is a maximum term number that a student may register as an undergraduate, denoted by $S_{max}$ which can be set to an arbitrarily large number, if desired, but which we set to 25 (5 terms per year, for a maximum allowed 5 years of study). A student does not have to register for courses in every such term: many students will only register for one or two courses for each of the Summer1 and/or Summer2 terms, but they will not register for any course offered in the SummerTerm term.

Finally, courses are only available during certain terms. The parameters $o_{i,s}$ take binary values to indicate whether course $c_i$ is offered in term $s \in \{1, \dots S_{max}\}$ with $o_{i,s} = 1$ meaning that course $c_i$ is offered in term $s$.

## 2.2 Problem Variables

We find it most useful to consider the following variables in order to formulate our problem:

- $x_i \in \{0,1\}, \forall i = 1, \dots N$ is 1 if student takes (and passes) course $c_i$ and 0 otherwise
- $x_{i,s} \in \{0,1\} \forall i = 1, \dots N, s = 0,1, \dots S_{max}$ is 1 if student takes (and passes) course $c_i$ in term $s$, and 0 otherwise. In particular, by $x_{i,0} = 1$ we denote courses the student transfers from other institutions, and in any way, courses the student passes before their first registration at the college. It is easy to see that $\sum_{s=0}^{S_{max}} x_{i,s} = x_i$ and therefore the entire sum will be either 1 or 0 according to whether or not the student ever takes and passes course $c_i$.
- $D \in \mathbb{N}$ denotes the last term number the student registers for any course before graduation. Obviously, it holds that $D \leq S_{max}$.
- $D_L \in \mathbb{N}$ denotes the maximum sum of difficulty levels of courses the student will encounter during any term. Obviously, any student would like this number to be kept as small as possible (so that they are not "overloaded" with difficult courses at any point during their studies).
- $G_e \in \mathbb{N}$ denotes the maximum sum of estimated grades the student will accumulate when taking the courses in the proposed plan. Obviously, any student would like this number to be as high as possible, so as to maximize their GPA upon graduation.

## 2.3 Problem Constraints

The problem constraints are as follows:

### 2.3.1 Pre-requisite and Co-requisite Constraints

Consider the directed graph of pre-requisite courses: every course $c_j$ has a (possibly empty) list of course pre-requisites $P_j = \{k | c_k < c_j\}$, where the inequality "$c_k < c_j$" is interpreted as "course $c_k$ must be completed before student registers for course $c_j$". These lists create the graph of course pre-requisites $G(V, E)$ where the nodes of the graph are the courses themselves, and there is a (directed) edge $(c_i, c_j) \in E$ if and only iff $i \in P_j$.

Further, consider the also directed graph of co-requisite courses: every course $c_j$ has a (possibly empty) list of course co-requisites $Q_j = \{k | c_k \leq c_j\}$, where the inequality "$c_k \leq c_j$" is interpreted as "course $c_k$ must either be completed before student registers for course $c_j$ or else be taken at the same term that student takes course $c_j$". These lists create the graph of course pre-requisites $G'(V, E')$ where the nodes of the graph are the courses themselves, and there is a (directed) edge $(c_i, c_j) \in E'$ if and only iff $i \in Q_j$.

Given the above definitions, we model the pre-requisite constraints for the scheduling of courses as follows:

$$x_{j,s} \leq \sum_{t=0}^{s-k_s} x_{i,t} \quad \forall (c_i, c_j) \in E, s = 1 \ldots S_{max} \text{ where } k_s = \begin{cases} 3 \text{ if } s\%5 == 0 \\ 1 \text{ else} \end{cases} \quad (1)$$

The meaning of the constraint is simple: $x_{j,s}$ cannot be 1 unless student has taken course $c_i$ at any term $t = 0, 1, \ldots s - k_s$ where $k_s = 1$ if $s$ is a "normal" term, and $k_s = 3$ if $s$ is a summer term, in which case, the student must have passed course $c_i$ at least by the previous Spring term.

**Extension to complex pre-requisite constraints**

At ACG (and the IT Department in particular), pre-requisite constraints can be a lot more complex than what was just described above. There are courses for which a student can register as long as they have completed one from a chosen set of other classes. As an example, for the course ITC4140 "Methods in ICT Project Research and Management", the student can register for it if they have successfully passed the course MA2010 "Statistics I" OR course MA2021 "Applied Statistics for Business" OR MA2025 "Applied Statistics for Science" (all, collectively known as "statistics courses"). In general, such a disjunctive constraint of prerequisite courses where course $c_{i_1}, c_{i_2}$ or $c_{i_3}$ must be completed before registering for course $c_j$ can be modeled via the following constraint:

$$x_{j,s} \leq \sum_{m=1}^{3} \sum_{t=0}^{s-k_s} x_{i_m,t} \quad s = 1 \ldots S_{max} \text{ where } k_s = \begin{cases} 3 \text{ if } s\%5 == 0 \\ 1 \text{ else} \end{cases} \quad (1')$$

Some courses however have even more complex prerequisite constraints: course ITC3287 "Advanced Object Oriented and Functional Programming" lists as its prerequisites the following: (1) ITC2088 "Introduction to Programming, and (2) ITC2197 "Object Oriented Programming Techniques" OR ITC3234 "Object Oriented Programming". This set of constraints is a CONJUNCTION of DISJUNCTIONS: the student must certainly have passed ITC2088, and must have taken at least one of ITC2197 and ITC3234. This conjunction is easily written down as the following 2 constraints:

$$x_{3287,s} \leq \sum_{t=0}^{s-k_s} x_{2197,t} \text{ where } k_s = \begin{cases} 3 \text{ if } s\%5 == 0 \\ 1 \text{ else} \end{cases}$$

$$x_{3287,s} \leq \sum_{t=0}^{s-k_s} (x_{2197,t} + x_{3234,t}) \text{ where } k_s = \begin{cases} 3 \text{ if } s\%5 == 0 \\ 1 \text{ else} \end{cases}$$

Such disjunctive constraints are particularly useful when a degree program undergoes changes and a course is "phased-out" in favor of another new course, but students can easily be caught in the midst of such a transition. In general, any set of pre-requisite constraints at ACG in its most general form is written as a disjunction of conjunctions (DNF) which can automatically be converted to a Conjunctive Normal Form (CNF) that is then modeled as a number of simultaneous constraints that must hold for each conjunct in the CNF, and each constraint is then of the form (1') above.

Similarly, the co-requisite constraints for the scheduling of the courses are as follows:

$$x_{j,s} \leq \sum_{t=0}^{s-k_s} x_{i,t} + x_{i,s} \quad \forall (c_i, c_j) \in E', s = 1 \ldots S_{max} \text{ where } k_s = \begin{cases} 3 \text{ if } s\%5 == 0 \\ 1 \text{ else} \end{cases} \quad (2)$$

Constraints (2) are essentially the same as (1), but now it is allowed for the student to take course $c_j$ in term $s$ as long as course $c_i$ has been taken until term $s - k_s$ or if $x_{i,s} = 1$ which means that course $c_i$ is taken concurrently with $c_j$ in the same term. There are no complex co-requisite constraints currently at ACG, even though it is easy to incorporate them as well if needed.

### 2.3.2 Level Constraints

Given the Level-k courses denoted by $L_4, L_5, L_6$ in section 2.1, there are some complex-sounding constraints that go as follows:

- For a student to register for a Level-5 course, they must have completed at least 4 Level-4 courses
- For a student to register for a Level-6 course, they must have completed ALL required Level-4 courses, and they must have completed at least 4 Level-5 courses

These constraints cannot be modeled in terms of variables indicating the term in which a student takes a course.

However, we can still model all Level-5 constraints using the binary $x_{i,s}$ variables as follows:

$$x_{i,s} \leq \frac{1}{4} \sum_{\substack{t=0 \\ j \in L_4}}^{s-k_s} x_{j,t} \quad \forall i \in L_5, \forall s = 1, \ldots S_{max} \text{ where } k_s = \begin{cases} 3 \text{ if } s\%5 == 0 \\ 1 \text{ else} \end{cases} \quad (3)$$

The meaning of constraints (3) is simple: until the term $s$ for which the variables $x_{j,t}$ sum to at least 4 between terms 0 and $s - k_s$, the constraint is active and since the RHS of the constraint is less than 1, it forces the binary variables $x_{i,s}$ to stay at zero, preventing the student from taking any Level-5 course in such terms; afterwards however, the constraint simply becomes inactive, allowing the student to take the course at this or any later term if they want.

Similarly, the Level-6 constraints are modeled as follows:

$$x_{i,s} \leq \frac{1}{4} \sum_{\substack{t=0 \\ j \in L_5}}^{s-k_s} x_{j,t} \quad \forall i \in L_6, \forall s = 1, \ldots S_{max} \text{ where } k_s = \begin{cases} 3 \text{ if } s\%5 == 0 \\ 1 \text{ else} \end{cases} \quad (4)$$

$$x_{i,s} \leq \frac{1}{|L_4|} \sum_{\substack{t=0 \\ j \in L_4}}^{s-k_s} x_{j,t} \quad \forall i \in L_6, \forall s = 1, \ldots S_{max} \text{ where } k_s = \begin{cases} 3 \text{ if } s\%5 == 0 \\ 1 \text{ else} \end{cases} \quad (5)$$

The constraints (4) ask for at least 4 Level-5 courses to be completed before a Level-6 class becomes available for a student; constraints (5) require that ALL Level-4 courses are completed before a Level-6 class becomes available to the student.

### 2.3.3 Class Availability Constraints

Of course, courses are not offered every term (see section 1.1). Therefore, the following constraints must hold to indicate that a student cannot register for a course at a time when the course is not offered by the college:

$$x_{i,s} \leq o_{i,s} \quad \forall i \in \{1, \ldots, N\}, s = 1, \ldots S_{max} \quad (6)$$

### 2.3.4 Credit and Subset Selection Constraints

As mentioned already, the student must accumulate a number of credits to graduate (see section 1.1):

$$\sum_{i=1}^{N} \bar{c}_i x_i \geq T_c \quad (7)$$

The value of the variable $x_i$ is simply the sum of the variables $x_{i,s}$ over all terms:

$$\sum_{s=0}^{S_{max}} x_{i,s} = x_i \ \forall i = 1, \ldots N \tag{8}$$

Of the total credits the student accumulates, there is a minimum required number of credits that must come from Liberal Education courses (see section 2.1):

$$\sum_{i \in LE} \bar{c}_i\, x_i \geq L_c \tag{9}$$

Every semester, the student can only take up to the maximum number of credits $C_{max}$ (see section 1.1):

$$\sum_{i=1}^{N} \bar{c}_i\, x_{i,s} \leq C_{max} \ \forall s = 1, \ldots S_{max} \tag{10}$$

Finally, for each set $R_j \subset C\ j = 1 \ldots k$, the student must take at least $r_j$ courses:

$$\sum_{i \in R_j} x_i \geq r_j \ \forall j = 1, \ldots k \tag{11}$$

The constraints (11) model the requirement that once a student chooses an "area of concentration" or "pathway", they must take at least a certain number of courses belonging to this concentration group.

### 2.3.5 Individual Course Selection and/or Assignment Constraints

A student may easily declare their intent to take a set of specific classes $\emptyset \subseteq S_d \subset C$. The constraint to do that is simply written as:

$$x_i = 1 \ \forall i \in S_d \tag{12}$$

In a similar fashion, one may wish to take a particular course $c_i$ on a particular term $t$. The constraint simply becomes

$$x_{i,t} = 1 \tag{13}$$

Notice that introducing the above constraints may make the optimization problem infeasible if any of the above constraints conflicts with the rest of the constraints 1.3.1--1.3.4.

### 2.3.6 Summer Terms Off Constraints

In case the student does not want to take any courses during the summer terms, this desire is modelled as follows:

$$x_{i,s} = 0 \ \forall i = 1, \ldots N, \forall s \in \{s_n, \ldots S_{max}\} \wedge (s\%5 \notin \{1,2\}) \tag{14}$$

where $s_n$ is the current term.

### 2.3.7 Soft-order Precedence Constraints

While working on this problem, we realized that in most programs there are courses the students take that would be better if they were taken in the opposite order from which the students often take them. For example, course ITC2095 "Software Engineering Practices" would be better understood by the student if the student had already taken ITC3160 "Fundamentals of RDBMS", yet neither course is pre-requisite or co-requisite for the other; in fact, ITC3160 is a higher level (Level-5) course whereas ITC2095 is a lower-level course (Level-4) and this level-order suggests to many students they should take them in the wrong order. To avoid such problems, we implement a soft-order precedence constraint between two courses $c_i, c_j$ to indicate that if a student takes both courses, then they should be taken in the order $c_i \rightarrow c_j$ ($c_i$ first, then $c_j$). This is modeled according to the inequalities (15) below:

$$x_{j,s} \leq \sum_{t=0}^{s-k_s} x_{i,t} + 1 - x_i \quad s = 1 \ldots S_{max} \text{ where } k_s = \begin{cases} 3 \text{ if } s\%5 == 0 \\ 1 \text{ else} \end{cases} \quad (15)$$

If the user does not take the course $c_i$ then $x_i = 0$ and the constraint becomes inactive; if however the user does take course $c_i$ then the constraint asks that if they take course $c_j$ that they take it on a term that follows the time they take $c_i$ by at least one term under "normal circumstances", and by at least 3 terms if the term they take $c_j$ is "SummerTerm" that runs concurrently with Summer1 and Summer2. Further, if the user has already passed course $c_j$ then the variable $x_{j,0} = 1$ and $x_{j,s} = 0 \, \forall s = 1, \ldots S_{max}$ and the constraints (15) are simply inactive regardless of whether the student takes in the future course $c_i$.

### 2.3.8 Further Optional Constraints

The student may optionally indicate a maximum number $T_{mx} > 0$ of courses they desire to register for during any term; this is in addition to the credit-imposed constraints on the number of courses a student can register for during any term (see section 2.1). This is enforced by the following set of constraints:

$$\sum_{i=1}^{N} x_{i,s} \leq T_{mx}, \quad \forall s = 1 \ldots S_{max} \quad (16)$$

The student may also optionally indicate a maximum number of courses $\Theta_{mx} \geq 1$ to register for during the same term that they register for their B.Sc. thesis, denoted by $c_\theta$. In a similar fashion as above, this is enforced by the following set of constraints:

$$\sum_{\substack{i=1 \\ i \neq \theta}}^{N} x_{i,s} \leq (\Theta_{mx} - 1)x_{\theta,s} + M(1 - x_{\theta,s}) \quad \forall s = 1, \ldots S_{max} \quad (17)$$

where $M$ denotes any number greater than or equal to the largest number of courses a student is allowed to take per term (the quantity $C_{max}$ is a valid candidate since every course is always worth at least 1 credit). The constraints (17) are easily understood when we consider what they represent in the two complementary cases when $x_{\theta,s} = 0$ or $x_{\theta,s} = 1$. In the 1st case, the constraint is $\sum_{\substack{i=1 \\ i \neq \theta}}^{N} x_{i,s} \leq M$ which is always inactive given the value of the quantity $M$. In the 2nd case on the other hand, the constraint $\sum_{\substack{i=1 \\ i \neq \theta}}^{N} x_{i,s} \leq \Theta_{mx} - 1$ is enforced which constrains the total number of other courses taken during the same term that the student takes course $c_\theta$ to less than $\Theta_{mx}$.

## 2.4 Objective Functions

### 2.4.1 Shortest Completion Time

A common objective from a student's perspective is to graduate as soon as possible! The objective function is then written as follows:

$$\min D$$

Subject to the extra constraints:

$$\sum_{s=1}^{S_{max}} s x_{i,s} \leq D \quad \forall i = 1, \ldots N \quad (18)$$

### 2.4.2 Difficulty Balance

Another frequent desire among students is to have a balanced class requirements (in terms of total difficulty levels of classes) between semesters. The relevant objective then is written as an objective to minimize the difference between the total difficulty in any two semesters

until graduation. Given that the current semester is $s_n$, the objective to minimize is therefore written as

$$\min_{s,s'=1,\ldots S_{max}} \left| \sum_{i=1}^{N} \bar{d}_i x_{i,s} - \sum_{i=1}^{N} \bar{d}_i x_{i,s'} \right|$$

The above objective function is written in terms of absolute values, but can be easily re-written in terms of a Linear Mixed Integer Programming Problem (MILP) with the help of newly introduced continuous variables $t_{s,s'}$ as follows:

$$\min D_L$$

Subject to the extra constraints:

$$D_L \geq \sum_{i=1}^{N} \bar{d}_i x_{i,s} - \sum_{i=1}^{N} \bar{d}_i x_{i,s'} \quad \forall s,s' = 1 \ldots S_{max} \tag{19}$$

$$D_L \geq \sum_{i=1}^{N} \bar{d}_i x_{i,s'} - \sum_{i=1}^{N} \bar{d}_i x_{i,s} \quad \forall s,s' = 1 \ldots S_{max} \tag{20}$$

### 2.4.3 Maximizing the Expected GPA via AI

Even though the above objectives are quite significant, from the student point of view, the most important objective for any schedule is to maximize their expected GPA as the cumulative grade point average is considered the ultimate metric of success in school, at least during Bachelor's studies. We would like therefore to be able to come up with a schedule of courses whereby the expected student GPA would be maximized given the courses we suggest. At first glance, one might think that based on the difficulty level of courses introduced earlier in the beginning of section 2.1, a schedule that minimizes the total difficulty of all classes taken should lead to higher GPA overall, but this ignores the facts that difficulty level is associated mostly with class failure rates rather than actual grades, and also that the objective associated with difficulty levels is to distribute difficult classes across terms in the schedule; so that the difficulty level may only help students improve their GPA in an *indirect* manner. Further, the difficulty level approach is a purely static approach that does not take into account individual student performance on classes taken so far.

Instead, having access to fully anonymized student records, we developed a Machine Learning/Data Mining inspired approach whereby *we extract from the dataset containing all student course records the set of all quantitative association rules that hold on the dataset with minimum user-specified support and confidence* that are in one of the following 3 forms:

$$grade(c_i) \geq v_i \rightarrow grade(c_j) \geq v_j$$

$$grade(c_i) \geq v_i \wedge grade(c_j) \geq v_j \rightarrow grade(c_k) \geq v_k$$

$$grade(c_i) \geq v_i \wedge grade(c_j) \geq v_j \wedge grade(c_k) \geq v_k \rightarrow grade(c_m) \geq v_m$$

We extract the above ruleset with minimal support threshold set to 0.5% and minimal confidence threshold set to 90% using QARMA (Christou, Amolochitis & Tan, 2018), (Christou, 2019) a highly parallel/distributed tool developed by one of us, that has given excellent results in various diverse applications so far (see Christou et al. (2022) for applications in Predictive Maintenance and the fashion industry). The above rules predict student success on not yet taken courses based on student success on already taken courses; intuitively this makes sense, as common experience says for example, that a student that obtained a very high grade on calculus is likely to get a very high grade on applied math, and similarly, someone who gets a high grade on an Operating Systems course is likely to also get a high grade on a High Performance Computing course. Given this ruleset, we estimate for each student their grade $\hat{g}_c$ they are going to get on a course $c$ as the highest value $v_{mx}$ predicted by any rule in the ruleset that has $c$ as its consequent that fires given the student's course taking history. If no rule fires for a particular course given the user's history (which obviously is going to happen for all courses given a freshman student that has passed no courses yet) we simply do not assign any predicted grade to this course.

Given the above, the objective to optimize becomes

$$\max \frac{\sum_{\substack{i=1 \\ i \in A}}^{N} \hat{g}_i x_i}{\|\{i \in A : x_i = 1\}\|}$$

where $A$ is the set of all courses open for the student to take for which we have an estimate $\hat{g}$ of the grade the student will get *that is above a pre-defined threshold* $g_{thres} > 0$, with the threshold being set to 2.5 for a final grade that ranges between 0 and 4. The reason for this threshold's existence is to avoid proposing courses to the student that we estimate they will not do well, just because we luck estimates for other courses. To bring the above objective in MILP format, we rewrite it as follows, considering that at any given time, the student must take a known number of credits (say $\hat{C}_{rem} > 0$) to graduate:

$$\max G_e$$

Subject to the extra constraints:

$G_e \leq \sum_{i \in A} \hat{g}_i x_i$ (21)

$\sum_{i \in A} \bar{c}_i x_i \leq \hat{C}_{rem}$ (22)

The above maximization objective together with the constraints (21)-(22) aim to maximize the expected GPA of the student, given their past record, and constitutes a completely personalized approach to planning for student success.

**Why Not Use Neural Networks?**

It is worth noting that the above described rule-based approach cannot be easily replaced with popular techniques such as (Deep) Neural Networks (Gutierrez-Rojas et al. 2022), for several reasons. Besides the fact that if we used Neural Network technology we would need to train as many networks as there are available courses for the student to choose from which would constitute a significant waste of resources, the main concern applying this technology has to do with the *nature* of the dataset itself: the dataset is very sparse in the sense that every student registers for up to 40 courses, while there is a total of more than 180 courses available to the student to choose from. This implies that if the grade of a student in a course is a feature, the cells of every row in the dataset would be *more than* 75% *empty* (containing null values), and of course, for freshman students, their corresponding rows would be almost 100% empty. Neural networks do not handle missing values well, even with the advent of imputation (Lin & Tsai, 2020) and GAN-based learning techniques (Goodfellow et al, 2014), and certainly not in datasets that are 75% or more sparse.

### 2.4.4 Prioritizing and Combining Objectives

The objectives developed and discussed in sections 2.4.1-2.4.3 are not mutually exclusive and in fact, given student priorities can be easily combined in one single objective, according to the linear objective function

$$C(D, D_L, G_e) = b_G G_e + b_D D + b_{DL} D_L \qquad (23)$$

Even though it is possible to ask the student to consider objectives separately, and create schedules for each different objective separately, and perform what-if scenarios, we have found it more convenient to simply ask the student to choose what they consider their most important criterion for course planning, and the remaining two objectives are prioritized according to a fixed order: when the student considers their most important objective to be their expected GPA, the variable $G_e$ is multiplied by the coefficient $b_G = -1000$ (the overall problem is always a minimization problem), the variable $D$ (see section 2.4.1) is multiplied by the coefficient $b_D = 100$, and the variable $D_L$ (see section 2.4.2) is multiplied by the coefficient $b_{DL} = 1$; otherwise, when the student considers their most important objective the time to complete their studies, the coefficient $b_D = 1000$, the coefficient $b_G = -100$ and the coefficient $b_{DL} = 1$ again.

## 3. System Implementation

The schedule optimization takes the form of a desktop application that accepts user-input, creates the model with the combined objective function (23), and subject to the constraints (1)-(22), produces an LP-format file, and then calls an optimization solver (currently GUROBI) to solve the problem specified in the lp file. The GUI is shown in Fig. 1, where we show the results of an optimization run for a freshman student, planning to take their first course in the Fall 2022 semester.

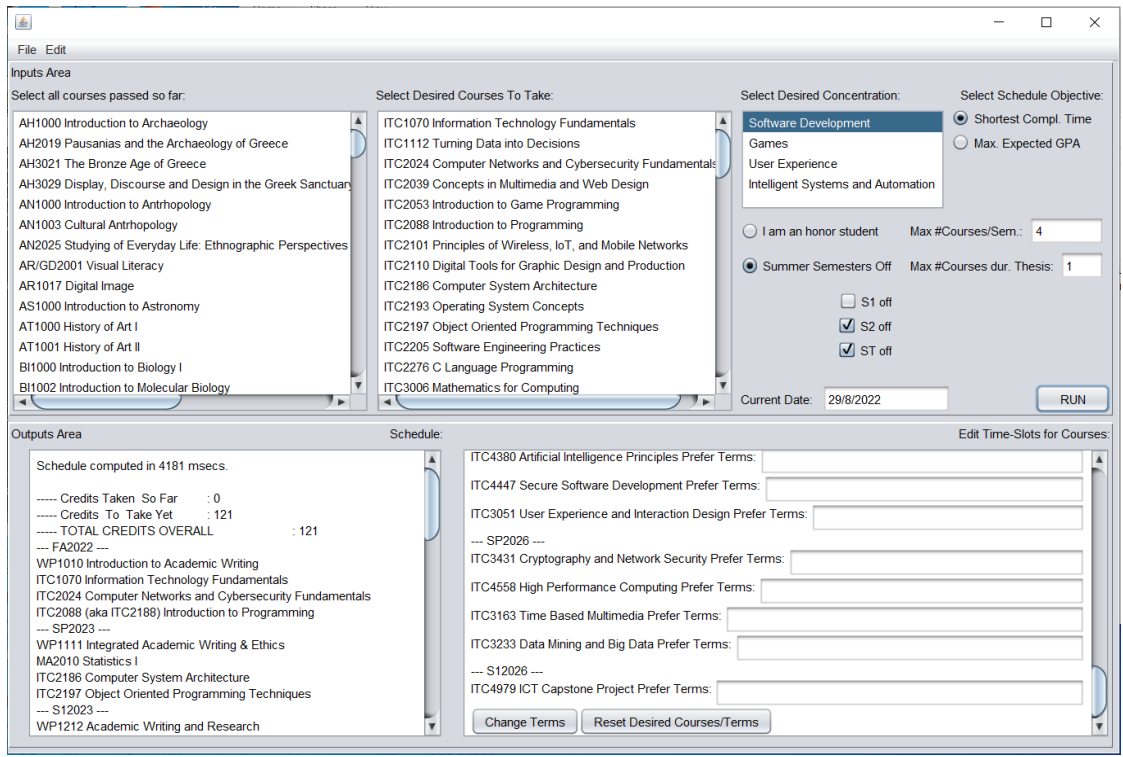

Figure 1: SCORER GUI

The idea is that the student must enter the courses they have passed so far (this can also be loaded from a file by choosing "Load" from the File menu), and enter the following information:

- Their desired concentration area which is used to determine a subset of courses that will be necessary to choose from,
- Any individual courses the student wishes to take from their program of study,
- The optimization objective to choose from, which sets priorities for all objectives, as discussed in 2.4.4,
- Whether the student is an honors student or not,
- Any student preferences towards registering for summer courses, and
- Student preferences towards maximum number of courses per term, and/or during their thesis

Given this information, by clicking "Run", the optimization problem is formulated, and the output when computed, written as text in the "Output Area" which can also be then saved in a text file. The outputs area, besides providing the full course schedule as computed by the optimizer's solution to the model, also includes an editor window in the lower right corner, under the heading "Edit Time-Slots for Courses" whereby the user can indicate their preferences towards the produced schedule; the user can indicate if they would like any proposed course to be scheduled at any other time, which they do so by indicating the terms when they would like to take the course; or they can simply reject the course by entering a minus sign "-" to indicate they don't want to take this course. Clicking the "Change Terms" button modifies the desired courses list area, allowing the next optimization run to take the new information into account.

## 3.1 System Design

As already mentioned, we developed the system as a small-scale (just above 5,000 loc) Java desktop application that uses the GUROBI Java API (Gurobi, 2022b) to communicate with the GUROBI optimizer, and the Java Swing API for GUI development. There is a small number of classes that together implement the system, as shown in the screenshot in Fig. 2.

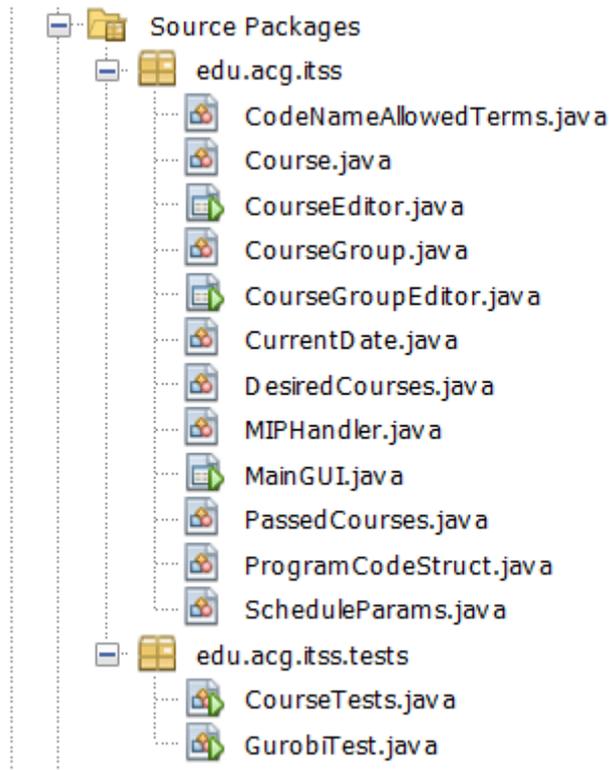

*Figure 2: ACG SCORER Modules/Class Design*

The heart of the system is the class `MIPHandler` which is responsible for reading all problem data, creating the relevant model, and passing it to the MIP solver to solve it. The classes `MainGUI`, `CourseEditor` and `CourseGroupEditor` implement the graphical user interfaces of the system.

### 3.2. Administrator Interfaces

To help department heads create the data files needed by our program, we also developed a couple of user interfaces. The first user interface guides an administrator edit course-related data (add/edit/delete courses), and is shown in fig. 3. In the same minimalistic design spirit that we used for our main GUI (fig. 1), this interface considers the screen to be a simple card displaying information about a single course at a time, and allowing the user to edit any information about the single course currently being displayed. If the user wants to save the changes they make to a course, they need to press the button "Save Course Changes", which saves the user edits, but only in memory (not to the disk). To save the relevant changes the user has made permanently, the user needs to press the button "Save to Disk", after having pressed "Save Course Changes".

Of course, a viable alternative for an administrator to edit courses, is to directly edit the relevant file (a standard ASCII CSV file) using a program such as Microsoft Excel.

*Figure 3: Course Editor GUI*

In a similar spirit to the design of our first administrator interface for editing courses, our second administrator interface helps authorized users create and edit course groups which are used in defining the constraints that any valid schedule must obey. The GUI is shown in fig. 4. The selected group "LE-core-stat" contains 3 courses (MA2010, MA2021, and MA2025, see discussion on extension to complex pre-requisite constraints in section 2.3.1) which are all statistics courses offered by the Math department. The fact that the check-box "#Courses denotes Maximum" is selected, means that the value in the text field "#Courses Constraint Value" (1) indicates the exact number of courses that should be taken from this group. The check-box "#Courses Constraint Applies Per Term" if checked, indicates that the constraint applies not globally, but on a semester level, essentially setting an upper limit on the number of courses from the group that the student is allowed to take during the same term. The user may edit the selected course group whose details appear in the GUI form at any given time by editing the group properties (possibly modifying even the courses that comprise the group) and then clicking on the "Save Group" button. New groups can be created using the "New Group" button, and existing groups can be deleted by selecting them from the list and pressing the "Remove Group" button.

Some constraints have special meaning that is assigned to them by the group's name. Therefore, any group whose name begins with the string "softorder" is a soft-order constraint, whose meaning we already discussed in section 2.3.7. Any group whose name begins with the string "HonorGroup" indicates classes only available to honor students. And the $L_4, L_5, L_6$ classes are contained in special course groups with the indicative names "L4", "L5" and "L6" respectively.

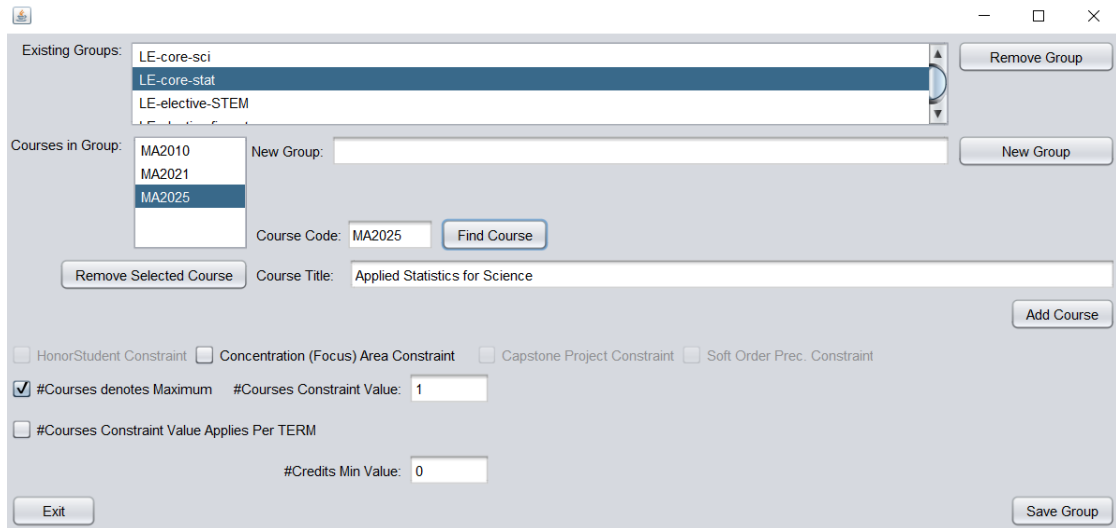

*Figure 4: Course Group (Constraints) Editor GUI*

## 4. Results

The Information Technology (IT) program at Deree offers a total of 69 courses in ITC whereas the total number of courses available to the students exceed 180. Scheduling a freshman who has yet to take their first class, results in a MIP with 3 continuous variables (the ones shown in the objective function) and 7854 binary variables and a total of 36,253 constraints. The Cyber-security and Networks (CYN) program is a separate program running at Deree that requires students to take 24 specific courses in that area, and another 4 courses in ITC, with the total number of courses available to students still exceeding 180. Such models are solved by GUROBI 9.5.1 on an old laptop equipped with a $3^{rd}$ generation intel core i-7 CPU with 16GB RAM usually in less than 10 seconds, making the program respond almost in real-time to the user requests, and thus allowing for extensive schedule editing and sensitivity analysis afterwards. Table 1 provides some data on the time it takes to produce useful schedules on such an old laptop for both the IT and CYN programs. As can be seen, the run-time never exceeds 3.5 seconds of wall-clock time; in fact, the more constrained is the student schedule (by courses already passed), the easier it becomes for the model to be optimized. The careful reader might notice that in fig. 1 which shows the results of the run corresponding to the 1st row of Table 1, the response time of the system is mentioned to have taken 4.18 seconds but this includes the time that it takes the system to gather the data, prepare and write the lp file, and call the GUROBI optimizer. The label "7+1" for the number of desired courses specified in the $3^{rd}$ row indicates that the student indicated 7 courses they would like to take, and 1 course they wish *not* to take.

*Table 1: ACG SCORER Running Times Under Various Scenarios*

| Program | #Credits Taken | #Desired Courses Specified | #Total Vars(Real) | #Constraints | Run-time (secs) |
|---|---|---|---|---|---|
| IT | 0 | 0 | 7857(3) | 36253 | 3.3 |
| | 0 | 5 | 7857(3) | 34762 | 3.2 |
| | 0 | 7+1 | 7857(3) | 34765 | 3.4 |
| | 30 | 0 | 7857(3) | 36253 | 2.2 |
| | 30 | 2 | 7857(3) | 36255 | 3 |
| | 49 | 0 | 7857(3) | 36253 | 2.1 |
| | 49 | 2 | 7857(3) | 34759 | 2.1 |
| | 79 | 0 | 7857(3) | 36253 | 1.1 |
| | 79 | 2 | 7857(3) | 36255 | 1.1 |
| | 91 | 0 | 7857(3) | 36253 | 1 |
| CYN | 0 | 0 | 7940(3) | 35466 | 3.5 |
| | 0 | 3 | 7940(3) | 35470 | 3.0 |
| | 30 | 0 | 7940(3) | 35466 | 3.4 |
| | 30 | 3 | 7940(3) | 35469 | 3.0 |
| | 60 | 0 | 7940(3) | 35466 | 1.5 |
| | 60 | 2 | 7940(3) | 35468 | 1.4 |
| | 90 | 0 | 7940(3) | 35466 | 1.3 |
| | 90 | 1 | 7940(3) | 35467 | 1.3 |

Given the consistent response times of the system, the department heads and program coordinators advising the undergraduate student body have reduced their load by more than 3 *orders of magnitude* (from around 3,600 seconds per student, to under 3.4 seconds!) What is equally, if not more, important, the system is able to suggest a plan of courses to the student that maximizes their expected GPA, which is a major Key Performance Indicator of measuring student success at the undergraduate level, something that was not previously possible to be done manually due to the enormous amount of effort required to comb through the data to estimate what courses would be a good fit for every student separately.

### 4.1. System Responsiveness under Open-Source MIP Solvers

As seen from the results of Table 1, the system responds very quickly to user commands, making it a highly interactive tool that can be used to edit schedules very fast, at least when the MIP problem is given to the state-of-the-art commercial solver GUROBI. In Table 2, we compare on the same machine (a laptop equipped with an intel core i7 3630QM CPU with 16GB RAM) run-times for schedule creation in the most difficult of circumstances, which is students having taken no credits so far, for the GUROBI 9.5 as well as SCIP 8.0 (https://www.scipopt.org), currently one of the fastest non-commercial, Open-Source MIP solvers in the world. As can be seen, SCIP is a viable alternative, always managing to find the optimal solution to the problem, but the run-time makes SCORER less interactive than it is when GUROBI is used as the underlying solver. In defense of Open-Source software, it can be argued that on a more modern machine, the running time of the SCIP solver should be expected to be reduced to much more acceptable levels (in the orders of *less than 1 minute for all cases.*)

Table 2: Comparing SCORER Run-Time Under Different MIP Solvers

| Program | Concentration Area | Solver Run-time (secs) | |
|---|---|---|---|
| | | GUROBI | SCIP |
| IT | S/W Development | 3.39 | 71 |
| | Games | 3.83 | 30 |
| | User Experience | 6.3 | 345 |
| | Intelligent Systems & Automation | 3.87 | 40 |
| CYN | CYN | 3.53 | 24 |

## 5. Conclusions and Future Directions

At ACG, the problem of advising students as to what classes to take in the course of their studies has been a manual, intensive labor process for the department heads and program coordinators, taking an estimated toll of up to 750 hours of human labor of faculty high up in the hierarchy every semester. Using modern advances in optimization software and modeling, we have managed to reduce this workload by offloading the problem to computers that can solve it almost instantaneously. The response times of the software, together with the course editing user interfaces we have implemented, allow students themselves to use the software and edit the resultant proposed schedules as many times as they wish until they get the course plan that feels exactly right for them.

What is more important, utilizing advanced data mining techniques, we have managed to be able to often get accurate estimates for the grade a student is going to get on a course, before they even register for it, an ability that we have since integrated into the proposed course plans that our automated software builds so as to maximize the student's expected GPA, the most significant measure of success during undergraduate studies.

The system is currently in use by the Information Technology and the Cyber-security and Networks programs at Deree College of the ACG. Our short-term plans include deploying our program for use by most if not all programs offered by the Deree College's undergraduate as well as graduate schools. We also plan to further improve the course grade predictor (currently based on QARMA) and test its limits, to see if it is possible to accurately estimate student success on a given program before the student is admitted to that program, while ensuring the absence of any biases in the system's decision making process. The explainable and interpretable nature of QARMA, as opposed to opaque methods such as neural networks, should offer significant help in this respect.